\documentclass[10pt,twocolumn,letterpaper]{article}

\usepackage{cvpr}
\usepackage{times}
\usepackage{epsfig}
\usepackage{graphicx}
\usepackage{amsmath}
\usepackage{amssymb}
\usepackage{multirow}
\usepackage{enumerate}
\usepackage{bm}

\usepackage[breaklinks=true,bookmarks=false]{hyperref}

\cvprfinalcopy 

\def\cvprPaperID{****} 

\begin{document}

\title{SeqXY2SeqZ: Structure Learning for 3D Shapes by Sequentially Predicting 1D Occupancy Segments From 2D Coordinates}

\author{Zhizhong Han$^{1,2}$, Guanhui Qiao$^{1}$, Yu-Shen Liu$^{1}$, Matthias Zwicker$^2$\\
$^1$School of Software, BNRist, Tsinghua University, Beijing, China \\
$^2$Department of Computer Science, University of Maryland, College Park, USA\\
{\tt\small h312h@umd.edu, qiaogh18@mails.tsinghua.edu.cn, liuyushen@tsinghua.edu.cn, zwicker@cs.umd.edu}
}

\maketitle

\begin{abstract}
   Structure learning for 3D shapes is vital for 3D computer vision. State-of-the-art methods show promising results by representing shapes using implicit functions in 3D that are learned using discriminative neural networks. However, learning implicit functions requires dense and irregular sampling in 3D space, which also makes the sampling methods affect the accuracy of shape reconstruction during test.
To avoid dense and irregular sampling in 3D, we propose to represent shapes using 2D functions, where the output of the function at each 2D location is a sequence of line segments inside the shape.
Our approach leverages 
the power of functional representations, but without the disadvantage of 3D sampling. Specifically, we use a voxel tubelization to represent a voxel grid as a set of tubes along any one of the X, Y, or Z axes. Each tube can be indexed by its 2D coordinates on the plane spanned by the other two axes. We further simplify each tube into a sequence of occupancy segments. Each occupancy segment consists of successive voxels occupied by the shape, which leads to a simple representation of its 1D start and end location. Given the 2D coordinates of the tube and a shape feature as condition, this representation enables us to learn 3D shape structures by sequentially predicting the start and end locations of each occupancy segment in the tube. We implement this approach using a Seq2Seq model with attention, called SeqXY2SeqZ, which learns the mapping from a sequence of 2D coordinates along two arbitrary axes to a sequence of 1D locations along the third axis. SeqXY2SeqZ not only benefits from the regularity of voxel grids in training and testing, but also achieves high memory efficiency. Our experiments show that SeqXY2SeqZ outperforms the state-of-the-art methods under widely used benchmarks.
\end{abstract}

\section{Introduction}
3D voxel grids are an attractive representation for 3D structure learning because they can represent shapes with arbitrary topology and they are well suited to convolutional neural network architectures.
However, these advantages are dramatically diminished by the disadvantage of cubic storage and computation complexity, which significantly affects the structure learning efficiency and accuracy of deep learning models.

Recently, implicit functions have been drawing research attention as a promising 3D representation to resolve this issue. By representing a 3D shape as a function, discriminative neural networks can be trained to learn the mapping from a 3D location to a label, which can either indicate the inside or outside of the shape~\cite{pifuSHNMKL19,chen2018implicit_decoder,MeschederNetworks} or a signed distance to the surface~\cite{xu2019disn,Park_2019_CVPR}. As a consequence, shape reconstruction requires sampling the function in 3D, where the 3D locations are required to be sampled near the 3D surface for training. Recent approaches based on implicit functions have shown superiority over point clouds in terms of geometry details, and advantages over meshes in terms of being able to represent arbitrary topologies. Although it is very memory efficient to learn implicit functions using discriminative models, these approaches require sampling dense 3D locations in a highly irregular manner during training, which also makes the sampling methods affect the accuracy of shape reconstruction during test.


To resolve this issue, we propose a method for 3D shape structure learning
by leveraging the advantages of learning shape representations based on continuous functions without requiring sampling in 3D.
Rather than regarding a voxel grid as a set of individual 3D voxels, which suffers from cubic complexity in learning, we represent voxel grids as functions over a 2D domain that map 2D locations to 1D voxel tubes.
This voxel tubelization regards a voxel grid as a set of tubes along any one of three dimensions, for example Z, and indexes each tube by its 2D location on the plane spanned by the other two dimensions, i.e., X and Y.
In addition, each tube is represented as a sequence of occupancy segments, where each segment consists of successive occupied voxels given by two 1D locations indicating the start and end points. Given a shape feature as a condition, this voxel tubelization enables us to propose a Seq2Seq model with attention as a discriminative model to predict each tube from its 2D location. Specifically, we leverage an RNN encoder to encode the 2D coordinates of a tube with a shape condition, and leverage an RNN decoder to sequentially predict the start and end locations of each occupancy segment in the tube. Because our approach essentially maps a coordinate sequence to another coordinate sequence, we call our method \textit{SeqXY2SeqZ}. Given the 2D coordinates of a tube, SeqXY2SeqY produces the 1D coordinates of the occupancy segments along the third dimension. Not only can SeqXY2SeqZ be evaluated with a number of RNN steps that is quadratic in the grid resolution during test, but it is also memory efficient enough to learn high resolution shape representations. 
Experimental results show that SeqXY2SeqZ outperforms the state-of-the-art methods. In summary, our contributions are as follows:

\begin{enumerate}[i)]
\item We propose a novel shape representation based on 2D functions that map 2D locations to sequences of 1D voxel tubes, avoiding the cubic complexity of voxel grids. Our representation enables 3D structure learning of voxel grids in a tube-by-tube manner via discriminative neural networks.
\item We propose SeqXY2SeqZ, an RNN-based Seq2Seq model with attention, to implement the mapping from 2D locations to 1D sequences. Given a 2D coordinate and a shape condition, SeqXY2SeqZ sequentially predicts occupancy segments in a 1D tube. It requires a number of RNN steps that grows only quadratically with resolution, and achieves high resolutions due to its memory efficiency.
\item SeqXY2SeqZ demonstrates the feasibility of generating 3D voxel grids using discriminative neural networks in a more efficient way, and achieves state-of-the-art results in shape reconstruction.
\end{enumerate}

\section{Related work}
Deep learning models have made big progress in 3D shape understanding tasks~\cite{Zhizhong2016b,Zhizhong2016,Han2017,HanTIP18,Zhizhong2018seq,Zhizhong2019seq,parts4features19,3DViewGraph19,3D2SeqViews19,HanCyber17a,l2g2019,p2seq18,MAPVAE19,wenxin_2020_CVPR,hutaoaaai2020,Hu2019Render4CompletionSM}. Recent 3D structure learning methods are also mainly based on deep learning models, while working on various 3D representations including voxel grids, point clouds, triangle meshes, and implicit functions.

\noindent\textbf{Voxel-based models. }Because of their regularity, many previous studies learned 3D structures from voxel grids with 3D supervision~\cite{ChoyXGCS16,conf/cvpr/Richter018} or 2D supervision with the help of differentiable renderers~\cite{YanNIPS2016,TulsianiZEM17,mvcTulsiani18,DBLP:conf/3dim/GadelhaMW17,WuNIPS2016,Gadelha2019}. Due to the cubic complexity of voxel grids, these generative models are limited to relatively low resolution, such as $32^3$. Recent studies~\cite{ChoyXGCS16,DBLP:conf/nips/0001WXSFT17,NIPS2018_7494} employed shallow 3D convolutional networks to reconstruct voxel grids in higher resolutions of $128^3$, however, the computational cost is still very large. To remedy this issue, some methods employed a multi-resolution strategy~\cite{DBLP:conf/3dim/HaneTM17,DBLP:conf/iccv/TatarchenkoDB17}. However, these methods are very complicated to implement and additionally require multiple passes over the input. Another alternative was introduced to represent 3D shapes using multiple depth images~\cite{conf/cvpr/Richter018}. However, it is hard to obtain consistency across multiple generated depth images during inference.

Different from these generative neural networks, we provide a novel perspective to benefit from the regularity of voxel grids but avoid their cubic complexity by leveraging discriminative neural networks in shape generation.

\noindent\textbf{Point cloud-based models. }As pioneers, PointNet~\cite{cvprpoint2017} and PointNet++~\cite{nipspoint17} enabled the learning of 3D structure from point clouds. Later, different variations have been proposed to improve the learning of 3D structures from 3D point clouds~\cite{FanSG17,p2seq18,MAPVAE19} or 2D images with various differentiable renderers~\cite{InsafutdinovD18,Navaneet2019,navaneet2019differ,Yifan:DSS:2019,lin2018learning}. Although point clouds are a compact and memory efficient 3D representation, they cannot express geometry details without additional non-trivial post-processing steps to generate meshes.

\noindent\textbf{Mesh-based models. }Meshes are also an attractive 3D representation in deep learning~\cite{WangZLFLJ18,Groueix_2018_CVPR,Chaowenpixel2019,KatoUH18,Liu:Paparazzi:2018,liu2018beyond,liu2019softras,DBLP:journals/corr/abs-1908-01210}. Supervised methods employed 3D meshes as supervision to train networks by minimizing the location error of vertices with geometry constraints~\cite{WangZLFLJ18,Groueix_2018_CVPR,Chaowenpixel2019}, while unsupervised methods relied on differentiable renderers to reconstruct meshes from multiple views~\cite{KatoUH18,Liu:Paparazzi:2018,liu2018beyond,liu2019softras,DBLP:journals/corr/abs-1908-01210}. However, these methods cannot generate arbitrary vertex topology but inherit the connectivity of the template mesh.

\noindent\textbf{Implicit function-based models. }Recently, implicit functions have become a promising 3D representation in deep learning models~\cite{pifuSHNMKL19,xu2019disn,MeschederNetworks,chen2018implicit_decoder,Oechsle_2019_CVPR,Park_2019_CVPR,DBLP:journals/corr/abs-1901-06802}. By representing a 3D shape as a 3D function, these methods employ discriminative neural networks to learn the function from a 3D location to an indicator labelling inside or outside of the shape~\cite{pifuSHNMKL19,chen2018implicit_decoder,MeschederNetworks} or a signed distance to the surface~\cite{xu2019disn,Park_2019_CVPR}. However, these methods required to sample points near 3D surfaces during training. To learn implicit functions without 3D supervision, Liu et al.~\cite{NIPS2019_Shichen} introduced a novel ray-based field probing technique to mine supervision from 2D images, similarly, a concurrent work~\cite{sitzmann2019srns} employed a network to bridge world coordinates to a feature representation of local scene properties. Although it is very memory efficient to learn 3D implicit functions using discriminative models in a point-by-point manner, it requires sampling dense and irregular 3D locations during training, which also makes the sampling methods affect the accuracy of shape reconstruction during test.

Although our method is also a discriminative network for 3D structure learning, it can benefit from the regularity of voxel grids by learning a 2D function. It is memory efficient and avoids the dense and irregular sampling during training.

\section{Overview}

\begin{figure}[htb]
  \centering
   \includegraphics[width=\linewidth]{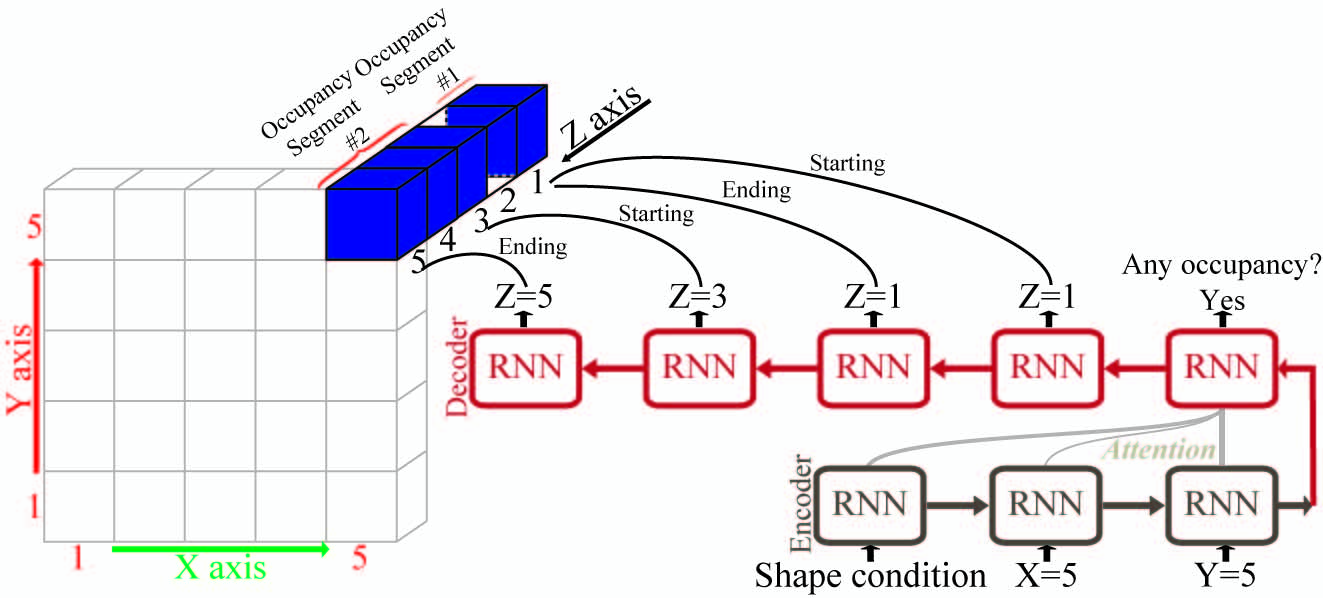}
  %
  %
\caption{\label{fig:frameworks} The overview of SeqXY2SeqZ. We interpret a voxel grids ($5\times5\times5$) as a set of tubes along any one (Z in our figure) of three axes, where each tube is indexed by 2D coordinates along the other two dimensions (X and Y in our figure). We represent each tube as a sequence of occupancy segments (shown by $\# 1$ and $\# 2$), and represent each occupancy segment using its 1D start and end location along the axis parallelling the tube. SeqXY2SeqZ generates a tube from its 2D coordinates (shown by X=5 and Y=5) and a shape condition by sequentially predicting the 1D start and end locations (shown by Z) of each occupancy segment in the tube. SeqXY2SeqZ consists of an RNN encoder and an RNN decoder with attention, where the RNN decoder also predicts whether there is any occupancy segment in the tube.}
\end{figure}

The core idea of SeqXY2SeqZ is to represent shapes as 2D functions that map each 2D location to a sequence of 1D occupancy segments. More specifically, we interpret each 3D shape $\bm{M}$ as a set of 1D tubes $\bm{t}_i$, where each tube $\bm{t}_i$ is indexed by its 2D coordinate $\bm{c}_i$. Tube $\bm{t}_i$ consists of a sequence of occupancy segments, where we represent each segment $o_j$ by its 1D start and end locations $s_j$ and $e_j$. To generate $\bm{M}$, SeqXY2SeqZ learns a 2D function to predict each tube $\bm{t}_i$ from its coordinate $\bm{c}_i$ and a shape condition by generating the start and end locations $s_j$ and $e_j$ of each occupancy segment $o_j$ in $\bm{t}_i$.


Fig.~\ref{fig:frameworks} illustrates how SeqXY2SeqZ generates a tube along the Z axis from its 2D coordinates on the X-Y plane. Specifically, we input the 2D coordinate $X=5$ and $Y=5$ sequentially into an encoder, and a decoder sequentially predicts the start and end locations of two occupancy segments along the Z axis. In the figure, there is one occupancy segment with only one voxel starting at $Z=1$ and ending at $Z=1$, and a second segment starting at $Z=3$ and ending at $Z=5$. Therefore, the decoder sequentially predicts $Z=1$, $Z=1$, $Z=3$, $Z=5$ to reconstruct the tube at $X=5$ and $Y=5$. In addition, the decoder outputs a binary flag to indicate whether there is any occupancy segment in this tube at all.
The encoder also requires a shape condition from an image or a learned feature as input to provide information about the reconstructed shape.

\section{Voxel Tubelization}

To train the SeqXY2SeqZ model, we first need to convert each 3D voxel grid into a tubelized representation consisting of sets of 1D voxel tubes over a 2D plane. For a 3D shape $\bm{M}$ represented by a grid with a resolution of $R^3$, voxel tubelization re-organizes these $R^3$ voxels into a set of $R\times R$ tubes $\bm{t}_i$ along one of the three axes. Each tube $\bm{t}_i$ can then be indexed by its location on the plane spanned by the other two dimensions using a 2D coordinate $\bm{c}_i$, such that $\bm{M}=\{(\bm{c}_i,\bm{t}_i)|i\in[1,R^2]\}$.
%
We further represent each tube $\bm{t}_i$ using run-length encoding of its $J_i$ occupancy segments $o_j$, where $j\in[1,J_i]$ and $J_i\in[1,R]$. An occupancy segment is a set of consecutive voxels that are occupied by the shape, which we encode as a sequence of start and end locations $s_j$ and $e_j$. Note that $s_j$ and $e_j$ are discrete 1D indices, which we will predict using a discriminative approach. We denote the tubes consisting of occupancy segments as $\bm{t}_i=[s_1,e_1,...,s_j,e_j,...,s_{J_i},e_{J_i}]$. In our experimental section we show that this representation is effective irrespective of the axis that is leveraged for the tubelization.
Our approach takes advantage of the following properties of voxel tubelization and run-length encoding of occupancy segments:

\begin{enumerate}[i)]
\item First, run-length encoding of occupancy segments significantly reduces the memory complexity of 3D grids, since only two indices are needed to encode each segment, irrespective of its length.


\item Second, our approach allows us to represent shapes as 2D functions that map 2D locations to sequences of 1D occupancy segments, which we will implement using discriminative neural networks. This is similar to shape representations based on 3D implicit functions implemented by discriminative networks, but our approach requires only $\mathcal{O}(R^2)$ RNN evaluation steps during shape reconstruction.

\item Third, networks that predict voxel occupancy using a scalar probability require an occupancy probability threshold as a hyperparameter, which can have a large influence on the reconstruction accuracy. In contrast, we predict start and end locations of occupancy segments and do not require such a parameter.

\end{enumerate}

\section{SeqXY2SeqZ}

SeqXY2SeqZ aims to learn to generate each tube $\bm{t}_i$ from its coordinate $\bm{c}_i$ and a shape condition. We use an RNN encoder to encode the coordinate $\bm{c}_i$ and the shape condition, while an RNN decoder produces the start and end locations of the occupancy segments $o_j$ in $\bm{t}_i$.

\noindent\textbf{RNN encoder. }
We condition the RNN encoder on a global shape feature $\bm{f}\in\mathbb{R}^{1\times D}$ that represents the unique 3D structure of each object.
For example, in 3D shape reconstruction from a single image, $\bm{f}$ could be a feature vector extracted from an image to guide the 3D shape reconstruction. In a 3D shape to 3D shape translation application, $\bm{f}$ could be a feature vector that can be jointly learned with other parameters in the networks, such as shape memories~\cite{Zhizhong2018VIP} or codes~\cite{Park_2019_CVPR}.

\begin{figure}[tb]
  \centering
   \includegraphics[width=\linewidth]{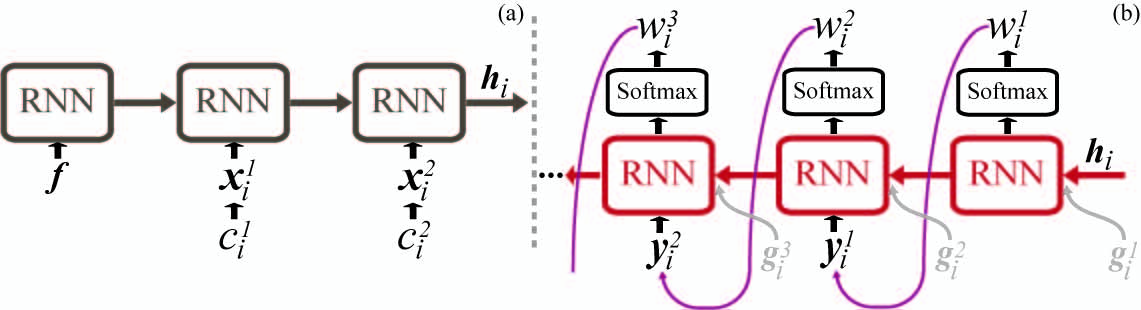}
  %
  %
\caption{\label{fig:rnn}(a) The RNN encoder encodes a shape condition $\bm{f}$ and the location embeddings $\bm{x}_i^1$ and $\bm{x}_i^2$ of 2D coordinate $\bm{c}_i=[c_i^1,c_i^2]$ indexing a tube into a hidden state $\bm{h}_i$. $\bm{h}_i$ can be further leveraged to generate the tube by the RNN decoder. (b) The RNN decoder is decoding the global occupancy indicator $b$ ($w_i^1$) and the start ($w_i^2$) and end ($w_i^3$) locations of the first occupancy segment in tube $\bm{t}_i$. The prediction at each step is produced by the hidden state at the previous step, context vector $\bm{g}_i^k$ by attention mechanism, and the embedding $\bm{y}_i^k$ of the location predicted at the previous step.}
\end{figure}

As shown in Fig.~\ref{fig:rnn}(a), the RNN encoder aggregates the shape condition $\bm{f}$ and a 2D coordinate $\bm{c}_i=[c_i^1,c_i^2]$ into a hidden state $\bm{h}_i$, which is subsequently leveraged by the RNN decoder to generate the corresponding tube $\bm{t}_i$. Rather than directly employing a location $c_i^1$ or $c_i^2$ as a discrete integer, we leverage the location as a location embedding $\bm{x}_i^1$ or $\bm{x}_i^2$, which makes locations meaningful in feature space. In this way, we have a location embedding matrix along each axis,  i.e., $\mathbf{F}_X$, $\mathbf{F}_Y$ and $\mathbf{F}_Z$. Each matrix holds the location embedding of all $R$ locations along an axis as $R$ rows, i.e., $\mathbf{F}_X\in\mathbb{R}^{R\times D}$, $\mathbf{F}_Y\in\mathbb{R}^{R\times D}$ and $\mathbf{F}_Z\in\mathbb{R}^{R\times D}$, so that we can get an embedding for a specific location by looking up the location embedding matrix. In the case of tubelizing along the Z axis demonstrated in Fig.~\ref{fig:frameworks}, the RNN encoder would employ the location embeddings along the X and Y axes, that is $\bm{x}_i^1=\mathbf{F}_X(c_i^1)$ and $\bm{x}_i^2=\mathbf{F}_Y(c_i^2)$.


We employ Gated Recurrent Units (GRU)~\cite{Cho2014On} as the RNN cells in SeqXY2SeqZ. At each step, a hidden state is produced, and the hidden state $\bm{h}_i$ at the last step is leveraged by the RNN decoder to predict a tube $\bm{t}_i$ for the reconstruction of a shape conditioned on $\bm{f}$, where $\bm{h}_i\in\mathbb{R}^{1\times H}$.

\noindent\textbf{Location embedding. }Although we could employ three different location embedding matrices to hold embeddings for locations along the X, Y, and Z axes separately, we use $\mathbf{F}_X$, $\mathbf{F}_Y$ and $\mathbf{F}_Z$ in a shareable manner. For example, we can employ the same location embedding matrix on the plane used for indexing the 1D tubes, such as $\mathbf{F}_X=\mathbf{F}_Y$ in the case shown in Fig.~\ref{fig:frameworks}. In our experiments, we justify that we can even employ only one location embedding matrix along all three axes, that is $\mathbf{F}_X=\mathbf{F}_Y=\mathbf{F}_Z$. The shareable location embeddings significantly increase the memory efficiency of SeqXY2SeqZ.

\noindent\textbf{RNN decoder. }With the hidden state $\bm{h}_i$ from the RNN encoder, the RNN decoder needs to generate a tube $\bm{t}_i$ for the shape indicated by condition $\bm{f}$ via sequentially predicting the start and end locations of each occupancy segment $o_j$.
To interpret the prediction of tubes with no occupancy segments, we include an additional global occupancy indicator $b$ that the decoder predicts first, where $b=1$ indicates that there are occupancy segments in the current tube.

We denote $\bm{w}_i$ as the concatenation of $b$ and $\bm{t}_i$, such that $\bm{w}_i=[b,s_1,e_1,...,s_{J_i},e_{J_i}]$, where each element in $\bm{w}_i$ is uniformly denoted as $w_i^k$ and $k\in[1,2\times J_i+1]$. Note that the start and end points $s_j$ and $e_j$ are discrete voxel locations, which we interpret as class labels. In each step, the RNN decoder selects a discrete label to determine either start or end location. Therefore, we leverage the following cross entropy classification loss to push the decoder to predict the correct label sequence $\bm{w}_i$ as accurately as possible under the training set,
\begin{equation}
L=-\sum_{k\in[1,2\times J_i+1]}\log p(w_i^k|w_i^{<k},\bm{h}_i),
\end{equation}
where $w_i^k$ is the $k$-th element in the sequence $\bm{w}_i$, $w_i^{<k}$ represents the elements in front of $w_i^k$, $p(w_i^k|w_i^{<k},\bm{h}_i)$ is the probability of correctly predicting the $k$-th element according to the previous elements $w_i^{<k}$ and the hidden state $\bm{h}_i$ from the encoder. Finally, our objective function is given as
\begin{equation}
\mathbf{F}_X^\ast,\mathbf{F}_Y^\ast,\mathbf{F}_Z^\ast,\bm{\theta}^\ast,\bm{f}^\ast=\min_{\mathbf{F}_X,\mathbf{F}_Y,\mathbf{F}_Z,\bm{\theta},\bm{f}} L,
\end{equation}
where $\bm{\theta}$ denotes the parameters of the RNN encoder and decoder, $\bm{f}$ is the shape condition, which is fixed or trainable depending on the application, and the location embedding matrices $\mathbf{F}_X,\mathbf{F}_Y,\mathbf{F}_Z$ could be shareable.


Training progress in a step by step manner is shown in Fig.~\ref{fig:rnn}(b). At the $k$-th step, element $w_i^k$ in sequence $\bm{w}_i$ is predicted through a softmax layer. For example, $w_i^1$ is either true or false for the global occupancy indicator $b$, and $w_i^2$ and $w_i^3$ are the start and end locations $s_1$ and $e_1$ of the occupancy segment $o_1$ in the range of $[1,R]$, etc. In addition, for each $w_i^k$ we look up its location embedding $\bm{y}_i^k$ from the location embedding matrix of the coordinate axis corresponding to the tube direction. The embedding $\bm{y}_i^k$ is then used in the prediction of $w_i^{k+1}$ at the next step. 
For example, in the tubelization along the Z axis demonstrated in Fig.~\ref{fig:frameworks}, $\bm{y}_i^k$ is looked up in $\bm{F}_Z$, such that $\bm{y}_i^k=\bm{F}_Z(w_i^k)$, where each row of $\bm{F}_Z$ represents an embedding for a location, and two additional rows for a true or false of $b$.

\noindent\textbf{Attention. }Finally, we leverage a state-of-the-art attention mechanism~\cite{BahdanauCB14} to increase the prediction accuracy of the predicted locations. We employ a context vector $\bm{g}_i^k$ for the prediction of $w_i^k$, where $\bm{g}_i^k$ summarizes how well each step of the encoder matches the prediction of $w_i^k$.

\section{Experiments and Analysis}


We employ tubelization along the Y axis in all our experiments and learn only two location embedding matrices. We share the location embedding matrices along the X and Z axes providing the 2D coordinates of tubes, such that $\mathbf{F}_X=\mathbf{F}_Z$, while we use a separate matrix along the Y axis. The location embedding is $D=512$-dimensional, and the hidden state of the RNNs is also $H=512$-dimensional, where the RNN encoder is bidirectional.


We train SeqXY2SeqZ using the Adam optimizer with $\epsilon=8\times10^{-6}$, with a batch size of $64$ and a learning rate of $1\times10^{-3}$ in all experiments. The maximum number of steps in the encoder and decoder are 4 and 30, respectively. We employ volumetric IoU to evaluate the accuracy of the reconstructed shapes, and all reported IoU values are multiplied by $10^2$.

\subsection{Representation ability}

\noindent\textbf{Dataset. }For fair comparison, we leverage five widely used categories from ShapeNetCore~\cite{ChangFGHHLSSSSX15} in this subsection, including airplane,
car, chair, rifle, and table, and keep the same train and test splitting as~\cite{chen2018implicit_decoder}. The ground truth shapes are also voxelized at a resolution of $64^3$, such that $R=64$.

\begin{table}[h]
\centering
\caption{The 3D reconstruction ($64^3$) comparison in terms of IoU.}  
\resizebox{\linewidth}{!}{
    \begin{tabular}{|c|c|c|c|c|c|}  
     \hline
         Methods & Plane & Car & Chair & Rifle & Table \\   
     \hline
       IM-AE~\cite{chen2018implicit_decoder}& 78.77  &	89.36 & 65.65 & 72.88& 71.44 \\
       CNN-AE~\cite{chen2018implicit_decoder}& 86.07 &	90.73 & 74.22 & 78.37& 84.67 \\
       OccNet(Train)~\cite{MeschederNetworks}& - &	- & 89.00 & -& - \\
       Our(512-512)&\textbf{90.35} &\textbf{91.18} & 74.32 & \textbf{84.46}& \textbf{86.21} \\
       Our(1024-2048)&- &- & \textbf{93.10} & -& - \\
     \hline
   \end{tabular}}
   \label{table:reconstruction}
\end{table}

\noindent\textbf{Auto-encoding. }We evaluate the representation ability of SeqXY2SeqZ in an auto-encoding task. We leverage a learnable shape condition $\bm{f}$ to represent each shape. Specifically, shape features $\bm{f}$ are learned together with the other parameters in the RNN during training. During testing, we keep updating the shape features while fixing the parameters in the RNN including the location embedding matrices, which is similar as  introduced by shape memories~\cite{Zhizhong2018VIP} or codes~\cite{Park_2019_CVPR}. Note that $\bm{f}$ are also $D=512$-dimensional vectors, similar as the location embeddings.

In this task, we compare SeqXY2SeqZ with results from the implicit decoder (IM)~\cite{chen2018implicit_decoder} and occupancy network (OccNet)~\cite{MeschederNetworks}. We show the comparison in Table~\ref{table:reconstruction}, where the mean IoU over the first 100 shapes in the test set of each category is reported by IM while OccNet only reported its results on the training set of chair at a resolution of 256.


As shown by ``Our(512-512)'' in Table~\ref{table:reconstruction}, our results with $D=512$-dimensional location embeddings and $H=512$-dimensional hidden states are the best among all compared methods under all shape categories. If we increase the learning ability of SeqXY2SeqZ by using location embeddings and hidden states with higher dimensions, such as $D=2048$ and $H=1024$ shown by ``Our(2048-1024)'', we achieve an even higher IoU of $93.10$ under the challenging chair class.

\begin{figure}[tb]
  \centering
   \includegraphics[width=\linewidth]{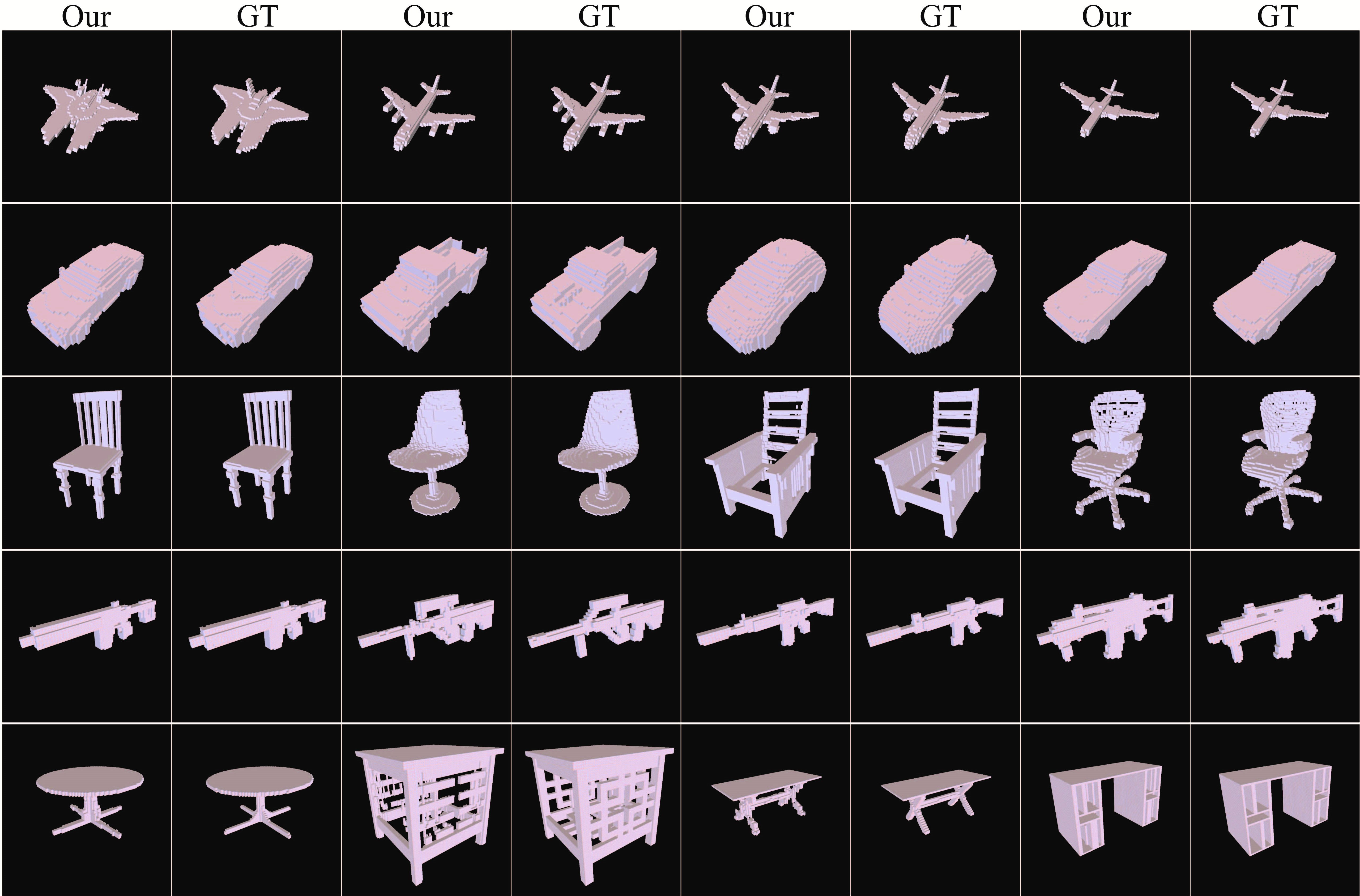}
  %
  %
\caption{\label{fig:3D23D} Auto-decoded shapes using learned 3D features.}
\end{figure}

In Fig.~\ref{fig:3D23D}, we visualize the reconstructed shapes in the test set of each category with our best results in Table.~\ref{table:reconstruction}. The reconstructed shapes with high fidelity demonstrate that SeqXY2SeqZ is capable of learning very complex structures of 3D shapes, such as the ones on chairs and tables.

\begin{table}[h]
\centering
\caption{Tubelization direction comparison.}  
    \begin{tabular}{|c|c|c|c|}  
     \hline
          & Along Y & Along Z & Along X  \\   
     \hline
       IoU& \textbf{90.35} & 89.96 & 90.21 \\
     \hline
   \end{tabular}
   \label{table:direction}
\end{table}

\begin{figure}[tb]
  \centering
   \includegraphics[width=\linewidth]{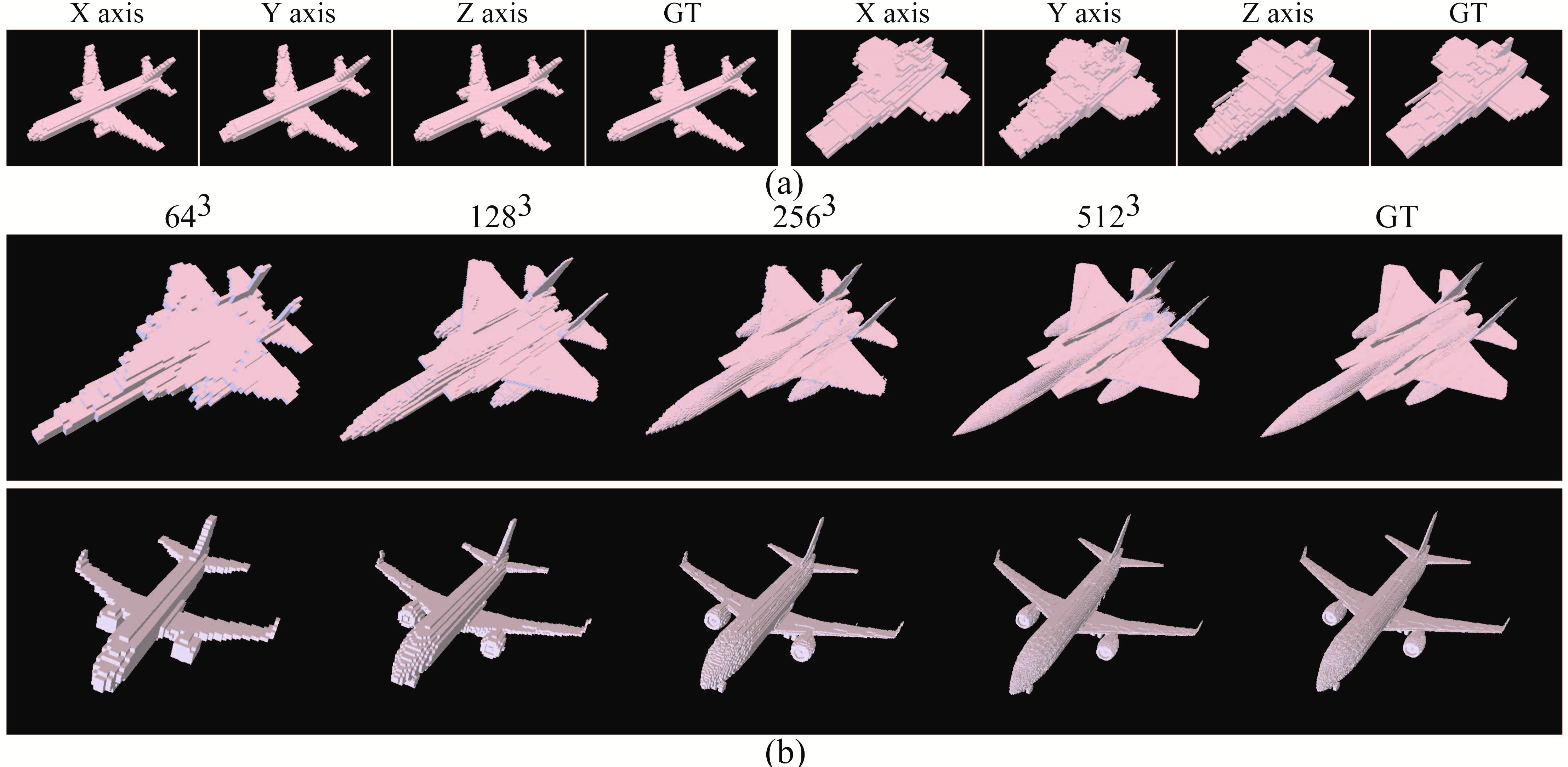}
  %
  %
\caption{\label{fig:direction}(a) Qualitative comparison of shapes reconstructed with tubelization along different axes. (b) Auto-encoded shapes in different resolutions.
}
\end{figure}

\noindent\textbf{Tubelization direction. }We can tubelize a voxel grid along any one of the X, Y or Z axes, which should be kept consistent in training and testing. Although the tubelization direction may lead to different ways of 3D structure learning, SeqXY2SeqZ does not exhibit any bias on the tubelization direction. We demonstrate this by training SeqXY2SeqZ using voxel grids tubelized under the X, Y and Z axis, respectively. Table~\ref{table:direction} shows that we achieve comparable results along the three tubelization directions under the airplane class. Visual comparisons are shown in Fig.~\ref{fig:direction}(a).

\noindent\textbf{High resolutions. }Thanks to the 2D functions and the shareable location embedding matrices, SeqXY2SeqZ is memory efficient enough to reconstruct shapes in high resolutions. We show auto-encoded airplanes in different resolutions in Fig.~\ref{fig:direction}(b). The high fidelity shapes justify our capabilities of high resolution reconstruction.

\subsection{Single Image 3D Reconstruction}

\noindent\textbf{Dataset. }We employ the dataset released from~\cite{ChoyXGCS16}, which contains 3D shapes from 13 categories in the ShapeNetCore~\cite{ChangFGHHLSSSSX15}. We also use the same train and test splitting, where each shape is represented as a voxel grid with a resolution of $32^3$ accompanying 24 rendered images. While many 3D reconstruction techniques (including ours, see Table~\ref{table:reconstruction} and Fig.~\ref{fig:direction}) support higher resolutions, we follow previous works~\cite{Tatarchenko_2019_CVPR,conf/cvpr/Richter018,liuaxivlihao,NIPS2019_Shichen} and choose ground truth voxel grids in the benchmark to provide a comparison to a broad range of competing approaches.


\noindent\textbf{Single image reconstruction. }We leverage a CNN encoder from~\cite{liu2019softras} to extract a 512 dimensional feature from a rendered image as a shape condition in this experiment. We compare with the state-of-the-art supervised and unsupervised methods in Table~\ref{table:reconstruction1}. Among these methods, ``DISN-V'' is a network formed by a DISN~\cite{xu2019disn} encoder and a 3D CNN decoder, ``DISN-C'' is DISN~\cite{xu2019disn} working with the estimated camera poses which is required in the reconstruction, ``PTN-R'' is the result using retrieval from PTN~\cite{YanNIPS2016}. Besides the voxel-based methods including R2N2~\cite{ChoyXGCS16}, PTN~\cite{YanNIPS2016} and Matryoshka~\cite{conf/cvpr/Richter018}, all the other methods represent 3D shapes as triangle meshes, where IM~\cite{chen2018implicit_decoder}, OccNet~\cite{MeschederNetworks}, and DISN~\cite{xu2019disn} are based on learning 3D implicit functions. For fair comparison, all the results listed here are taken from the literature rather than being reproduced by us. For example, the results of NMR~\cite{KatoUH18}, SoftRas~\cite{liuaxivlihao} and DIB-R~\cite{DBLP:journals/corr/abs-1908-01210} are all from DIB-R~\cite{DBLP:journals/corr/abs-1908-01210}.

\begin{figure*}[htb]
  \centering
   \includegraphics[width=\linewidth]{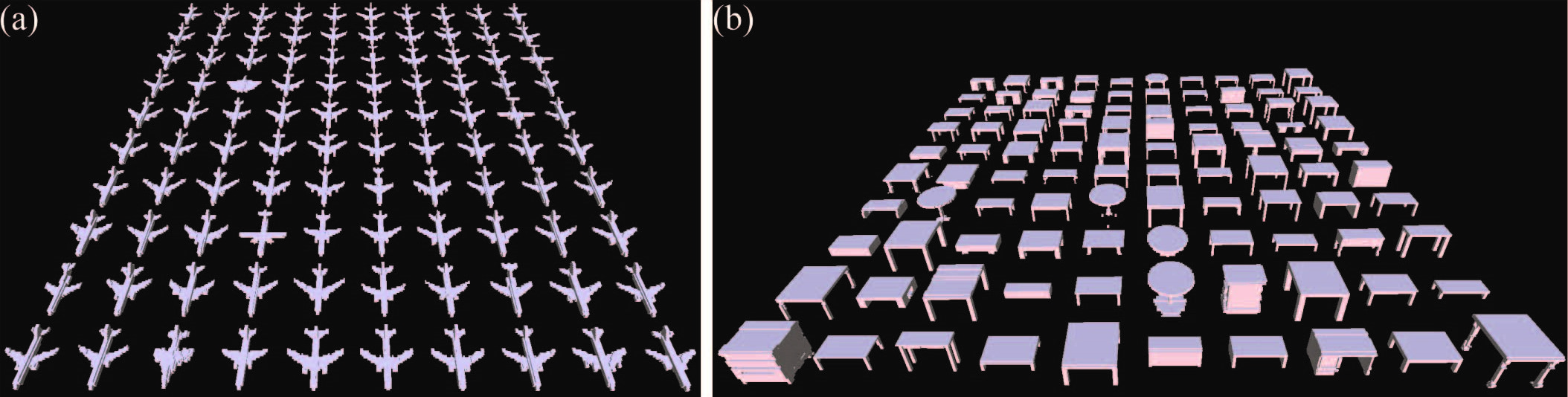}
  %
  %

\caption{\label{fig:2D23DVis} Single image reconstruction for airplanes in (a) and tables in (b).}
\end{figure*}
\begin{figure}[htb]

  \centering
   \includegraphics[width=0.95\linewidth]{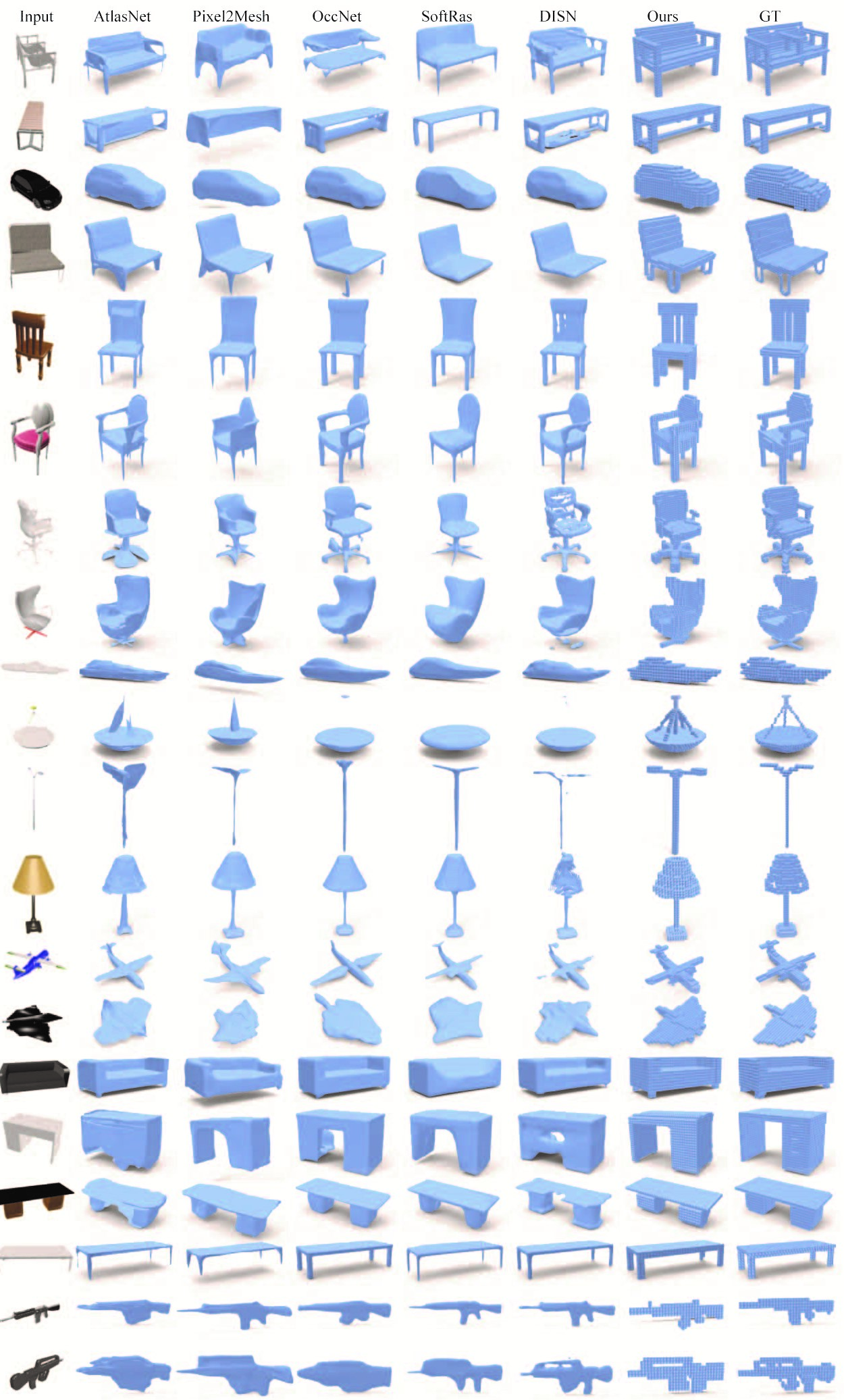}
  %
  %

\caption{\label{fig:compwithother}Qualitative comparison with the state-of-the-art supervised and unsupervised methods.
}
\end{figure}

Table~\ref{table:reconstruction1} demonstrates the performance of our method, showing that in terms of the mean IoU we improve by $6.3$ over the best 3D implicit function based method (DISN) and by $2.1$ over the best unsupervised method (DIB-R). We achieve the best IoU in 7 out of 13 categories among all supervised methods, and in 8 out of 13 categories among all unsupervised methods. Matryoshka~\cite{conf/cvpr/Richter018} comes closest to our performance, but it employs non-standard augmentation on training images, which we omit.
Fig.~\ref{fig:compwithother} shows a visual comparison, where the shapes are reconstructed from the same input images using the trained network parameters released by different methods. Although we trained our method at a resolution of $32^3$, the high accuracy enables us to reveal complex geometry that other methods cannot handle, which makes our results comparable to the meshes reconstructed by other methods. Fig.~\ref{fig:2D23DVis} shows additional airplanes and tables reconstructed by our method.

\begin{table*}[t]
\centering
\caption{Quantitative comparison of single image 3D shape reconstruction in terms of IoU.}  
\resizebox{\linewidth}{!}{
    \begin{tabular}{c|c|c|ccccccccccccc|c}  
     \hline
        &Method &Modality& Plane & Bench & Cabinet & Car & Chair & Display & Lamp & Speaker & Rifle & Sofa & Table & Phone & Boat & Mean  \\   
     \hline
       \multirow{10}{*}{\rotatebox{90}{Supervised}}&AtlasNet~\cite{Groueix_2018_CVPR}&\multirow{3}{*}{Mesh}&39.2&34.2&20.7&22.0&25.7&36.4&21.3&23.2&45.3&27.9&23.3&42.5&28.1&30.0\\
       &Pixel2mesh~\cite{WangZLFLJ18}&&51.5&40.7&43.4&50.1&40.2&55.9&29.1&52.3&50.9&60.0&31.2&69.4&40.1&47.3\\
       &3DN~\cite{wang20193dn}&&54.3&39.8&49.4&59.4&34.4&47.2&35.4&45.3&57.6&60.7&31.3&71.4&46.4&48.7\\
     \cline{2-17}
       &R2N2~\cite{ChoyXGCS16}&\multirow{2}{*}{Voxel}&51.3&42.1&71.6&79.8&46.6&46.8&38.1&66.2&54.4&62.8&51.3&66.1&51.3&56.0\\    &Matryoshka~\cite{conf/cvpr/Richter018}&&64.7&57.7&\textbf{77.6}&\textbf{85.0}&\textbf{54.7}&53.2&40.8&\textbf{70.1}&61.6&68.1&\textbf{57.3}&75.6&\textbf{59.1}&63.5\\
     \cline{2-17}
       &IM~\cite{chen2018implicit_decoder}&\multirow{4}{*}{3D Implicit}&55.4&49.5&51.5&74.5&52.2&56.2&29.6&52.6&52.3&64.1&45.0&70.9&56.6&54.6\\
       &OccNet~\cite{MeschederNetworks}&&54.7&45.2&73.2&73.1&50.2&47.9&37.0&65.3&45.8&67.1&50.6&70.9&52.1&56.4\\
       &DISN-V~\cite{xu2019disn}&&50.6&44.3&52.3&76.9&52.6&51.5&36.2&58.0&50.5&67.2&50.3&70.9&57.4&55.3\\
       &DISN-C~\cite{xu2019disn}&&57.5&52.9&52.3&74.3&54.3&56.4&34.7&54.9&59.2&65.9&47.9&72.9&55.9&57.0\\
      \cline{2-17}
       &Ours&2D Implicit&\textbf{73.2}&\textbf{58.5}&71.0&78.1&50.3&\textbf{60.0}&\textbf{44.7}&62.2&\textbf{66.7}&\textbf{68.4}&55.0&\textbf{80.2}&58.4&\textbf{63.6}\\
       \hline
       \hline
       \multirow{6}{*}{\rotatebox{90}{Unsupervised}}    &NMR~\cite{KatoUH18}&\multirow{3}{*}{Mesh}&58.5&45.7&74.1&71.3&41.4&55.5&36.7&67.4&55.7&60.2&39.1&76.2&59.4&57.0\\
       &SoftRas~\cite{liuaxivlihao}&&58.4&44.9&73.6&77.1&49.7&54.7&39.1&68.4&62.0&63.6&45.3&75.5&58.9&59.3\\
       &DIB-R~\cite{DBLP:journals/corr/abs-1908-01210}&&57.0&49.8&\textbf{76.3}&\textbf{78.8}&52.7&58.8&40.3&\textbf{72.6}&56.1&67.7&50.8&74.3&\textbf{60.9}&61.2\\
       \cline{2-17}
       &PTN-R~\cite{YanNIPS2016}&\multirow{2}{*}{Voxel}&55.6&48.8&57.1&65.2&35.1&39.6&29.1&46.0&51.3&53.1&31.0&67.0&40.8&47.7\\
       &PTN~\cite{YanNIPS2016}&&55.6&49.2&68.2&71.2&44.9&54.0&42.2&58.7&59.9&62.2&49.4&75.0&55.1&57.4\\   \cline{2-17}
       &IMRender~\cite{NIPS2019_Shichen}&\multirow{1}{*}{3D Implicit}&65.1&53.6&-&78.2&\textbf{54.8}&-&-&-&-&-&51.5&-&60.8&60.7\\
       \cline{2-17}
       &Ours&2D Implicit&\textbf{73.2}&\textbf{58.5}&71.0&78.1&50.3&\textbf{60.0}&\textbf{44.7}&62.2&\textbf{66.7}&\textbf{68.4}&\textbf{55.0}&\textbf{80.2}&58.4&\textbf{63.6}\\
     \hline
   \end{tabular}}
   \label{table:reconstruction1}
 \end{table*}

\subsection{Ablation Studies and Analysis}

\noindent\textbf{Ablation studies. }We highlight some elements in our method by ablation studies in single image reconstruction under the chair class in Table~\ref{table:ablation}. We compare our result with the ones without attention (``NoAtt''), the ones with LSTM RNN cells (``LSTM''), and the ones with single direction RNN encoder (``SingleDir''). We find that GRU performs better than LSTM, and both attention mechanism and bidirectional RNN encoder contribute to the performance.

\begin{table}[h]
\centering
\caption{Ablation studies under chair class.}  
\resizebox{\linewidth}{!}{
    \begin{tabular}{|c|c|c|c|c|c|}  
     \hline
          & NoAtt &LSTM& SingleDir& ShareableXYZ & Our(GRU) \\   
     \hline
       IoU& 47.5 & 49.8 & 48.8&49.1 & \textbf{50.3} \\
     \hline
   \end{tabular}}
   \label{table:ablation}
\end{table}


\noindent\textbf{Shareable location embedding matrix. }The memory efficiency is one advantage of SeqXY2SeqZ. We achieve this not only by avoiding the direct involvement of 3D voxel grids, but also by sharing the location embedding matrices. The above experiments have shown the effectiveness of shared location embedding matrices for the X and Z axes to define the plane indexing the tubes. In this experiment, we step further by employing only one location embedding matrix for all three axes. We also tubelize the voxel grids along the Y axis, and train SeqXY2SeqZ under the chair class in sigle image 3D reconstruction. In Table~\ref{table:ablation}, ``ShareableXYZ'' still achieves the comparable result with ``Our(GRU)''.

\begin{figure}[h]
  \centering
   \includegraphics[width=\linewidth]{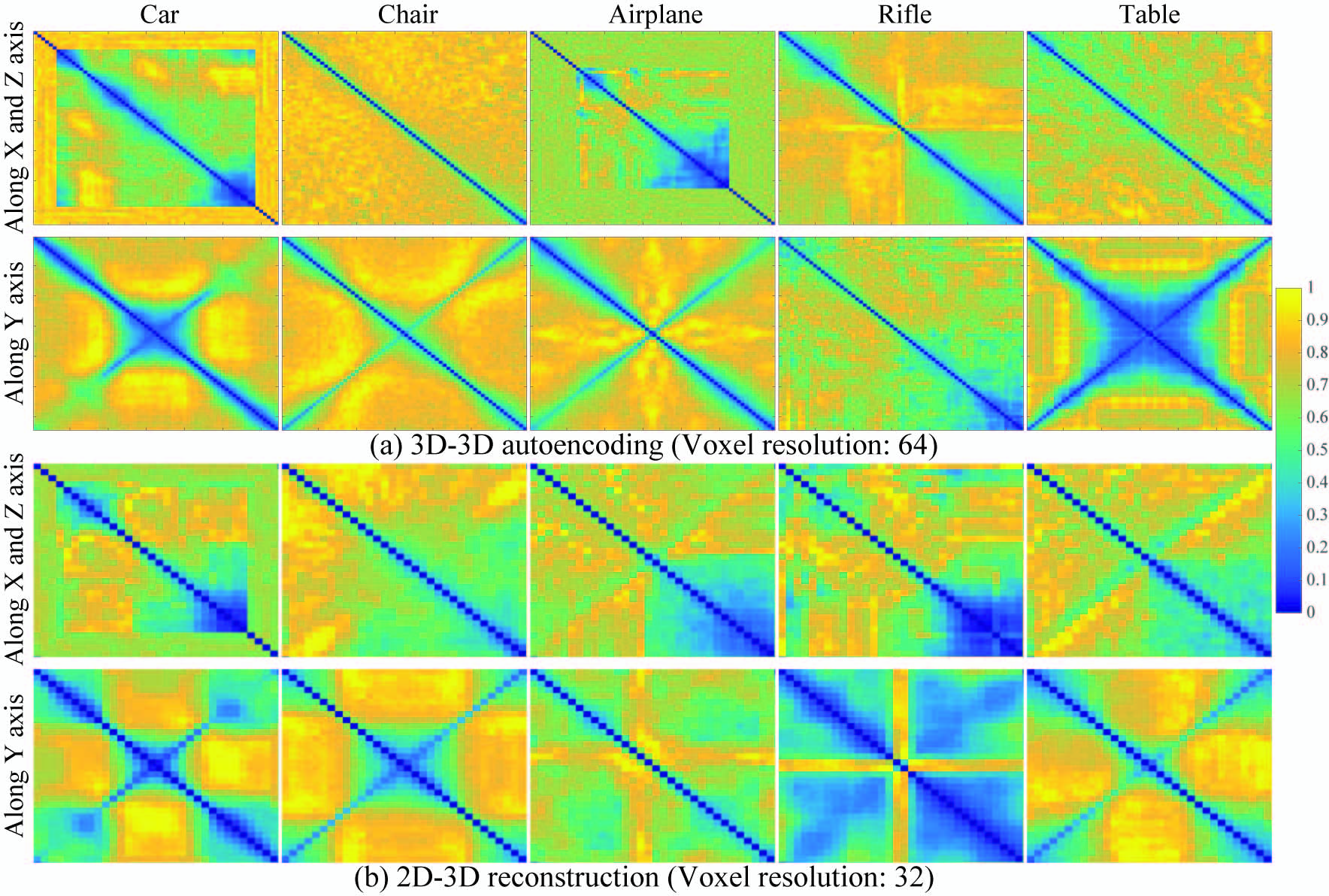}
  %
  %

\caption{\label{fig:embeddingvis} Pairwise cosine distances of location embeddings learned in auto-encoding (a) and single image reconstruction (b). In each similarity matrix, blue means two locations indicated by their embeddings show similar occupancy surrounding while yellow means more different.
}
\end{figure}

\noindent\textbf{Location embedding visualization. }We visualize the location embeddings learned in auto-encoding of Table~\ref{table:reconstruction} in Fig.~\ref{fig:embeddingvis} (a), where each class leverages two sets of location embeddings including one shared by the X and Z axes, and the other along the Y axis. We visualize each set of location embeddings using a cosine distance matrix whose element is the pairwise cosine distance between arbitrary two location embeddings.
The structure of a shape category is demonstrated by the distinctive patterns on the cosine distance matrix in different shape categories, which demonstrates the effectiveness of the learned location embeddings. In each similarity matrix, blue means more similar between two location embeddings while yellow means more different. The similarity indicates whether the two corresponding locations show similar occupancy surrounding. For a class containing shapes with similar structures, like cars, the patterns are more obvious, while a class containing shapes with large structure variations, like chairs, the patterns are less obvious. In addition, we visualize the location embeddings learned in single image reconstruction of Table~\ref{table:reconstruction1} in Fig.~\ref{fig:embeddingvis} (b), where we also observe the different patterns on the cosine distance matrix in different shape categories. Note that we show the $64^2$ dimensional distance matrix in Fig.~\ref{fig:embeddingvis} (a) and the $32^2$ dimensional distance matrix in Fig.~\ref{fig:embeddingvis} (b) in the same size.

\begin{figure}[htb]
  \centering
   \includegraphics[width=\linewidth]{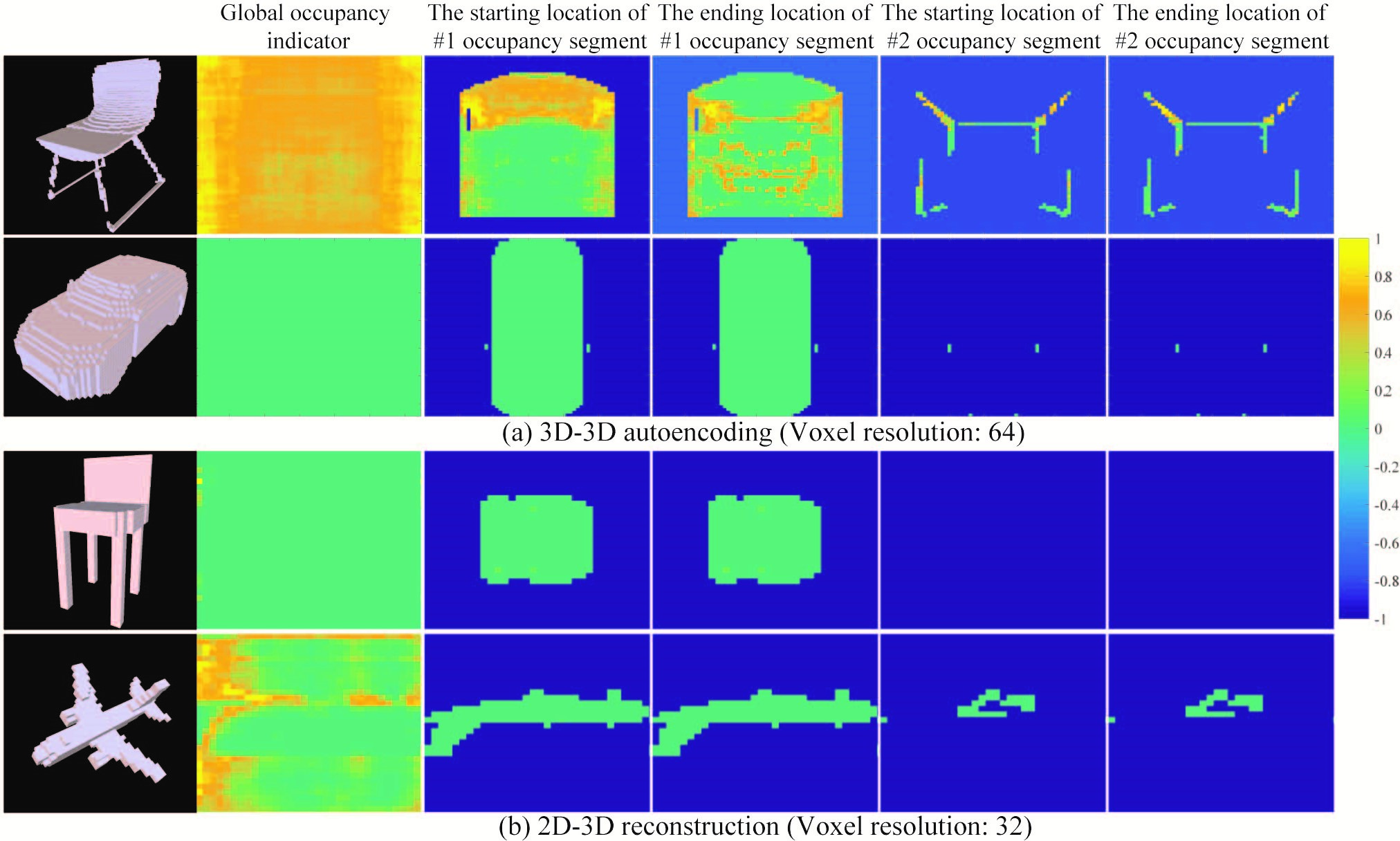}
  %
  %

\caption{\label{fig:attentionvis} The visualization of attention learned in auto-encoding. We visualize the attention weights learned at the first five steps of the decoder. The attention at each step for all 2D coordinates is shown as an image, where attention weights on the encoder at each 2D coordinate are encoded as entropy shown by color.
}
\end{figure}

\noindent\textbf{Attention visualization. }We further visualize the attention learned in auto-encoding of Table~\ref{table:reconstruction}. At each 2D coordinate, an attention vector $\bm{a}$ is learned at each decoder step for all encoder steps. For each decoder step, we leverage entropy ($-\bm{a}.*log_2\bm{a}$) to visualize $\bm{a}$ at all 2D coordinates (if there is no output at this decoder step, we encode $-1$ at this 2D coordinate) into an attention image, and we normalize the whole attention image using the maximal entropy. We show five attention images at the first five decoder steps for each shape in Fig.~\ref{fig:attentionvis} (a). In each image, the higher entropy (above 0, the lighter color) indicates this decoder step is paying attention more equally on all encoder steps to generate more complex structure, such as chairs, while the lower entropy (above 0, the darker color) indicates this decoder step is focusing on a specific encoder step to generate relatively simple structure, such as cars. Similarly, we visualize the attention learned in single image reconstruction of Table~\ref{table:reconstruction1} in Fig.~\ref{fig:attentionvis} (b), where the chair can be reconstructed by only one occupancy segment at all 2D coordinates, which makes the attention much simpler than the one for the chair in Fig.~\ref{fig:attentionvis} (a). Note that we show the $64^2$ dimensional attention images in Fig.~\ref{fig:attentionvis} (a) and the $32^2$ dimensional attention images in Fig.~\ref{fig:attentionvis} (b) in the same size.

\begin{figure}[htb]
  \centering
   \includegraphics[width=\linewidth]{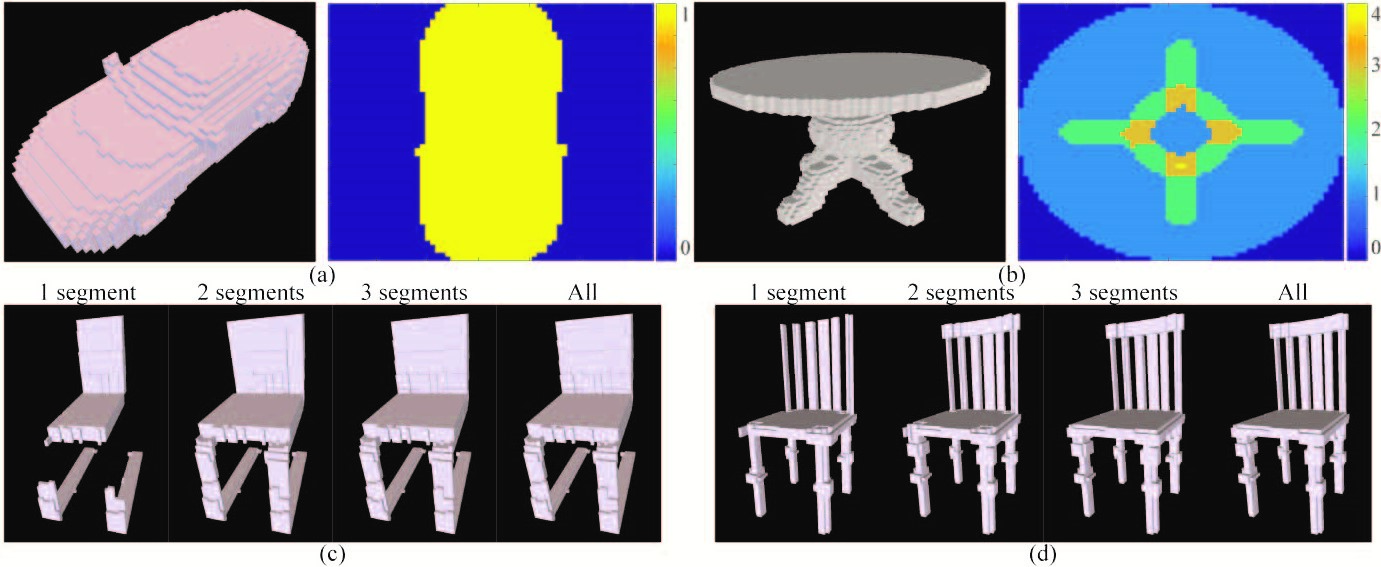}
  %
  %
\caption{\label{fig:OccSegNumvis} The efficiency demonstration of occupancy segments at 2D coordinates. We show the numbers of occupancy segments at all 2D coordinates as an image for a car (a) and table (b). We also visualize the shapes reconstructed by different numbers of occupancy segments at all 2D coordinates in (c) and (d). In all cases, only few occupancy segments are needed to represent a complex shape. }
\end{figure}

\noindent\textbf{Occupancy segment visualization. }With the auto-encoded shapes, we justify the efficiency of our voxel tubelization by visualizing the number of predicted occupancy segments at each 2D coordinate in Fig.~\ref{fig:OccSegNumvis}. For a simple car in Fig.~\ref{fig:OccSegNumvis} (a), it is enough to use only one occupancy segment to represent the geometry at each 2D coordinate. Although the table in Fig.~\ref{fig:OccSegNumvis} (b) is more complex, we can still leverage no more than three occupancy segments to represent the geometry at almost all 2D coordinates. In Fig.~\ref{fig:OccSegNumvis} (c) and (d), we visualize the reconstructed chairs with different numbers of occupancy segments. For complex structures in chairs, it is still enough to reconstruct almost a whole shape using only two occupancy segments.

\begin{figure}[htb]
  \centering
   \includegraphics[width=\linewidth]{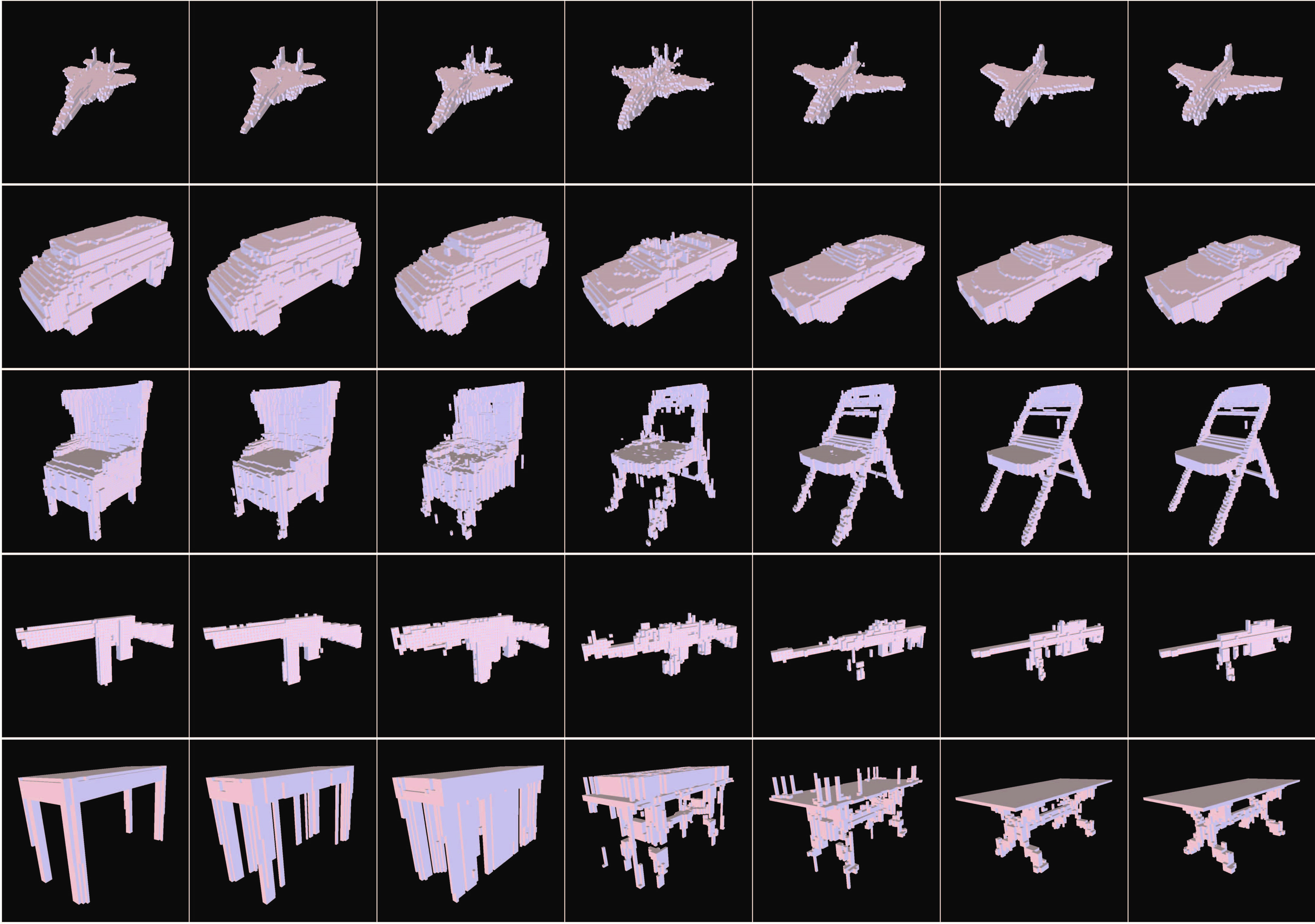}
  %
  %
\caption{\label{fig:Interpolation} Interpolated shapes in the learned 3D feature space. We use the learned shape conditions of two shapes to establish a line in the feature space, and then uniformly sample five features along this line by interpolating the two shape conditions. Finally, we generate a interpolated shape using each one of the five interpolated features as shape condition.}
\end{figure}

\noindent\textbf{Interpolation. }We visualize the shape condition space learned in auto-encoding of Table~\ref{table:reconstruction} by visualizing the interpolation between two shapes. We interpolate the features between two learned shape conditions, which are further leveraged to reconstruct shapes shown in Fig.~\ref{fig:Interpolation}. The transition shows how one shape is gradually transformed to another one by manipulating occupancy segments.

\noindent\textbf{Memory and Computation Time. }We compare the memory and computation time requirements with methods based on learning 3D implicit functions in Table~\ref{table:time}, including OccNet~\cite{MeschederNetworks} and DISN~\cite{xu2019disn}. To reconstruct a 3D shape at a resolution of $R^3$ from a single image during test, OccNet~\cite{MeschederNetworks} requires to get occupancy values for about $3.8*R^3$ sampled points with $sub$ additional steps of subdivision, while DISN requires to get SDF values for $R^3$ sampled points, both of which are higher complexity than our $\mathcal{O}(R^2)$ RNN steps. Since DISN cannot run on a single GPU as OccNet and SeqXY2SeqZ, we report a fair comparison in terms of the CPU run time and RAM space with $R=64$ and $sub=2$ for reconstructing one shape from a single image. Benefiting from learning 2D functions that predict sparse representations of 1D voxel tubes, SeqXY2SeqZ achieves both the lowest time and memory requirements by a large margin.

\begin{table}[h]
\centering
	\caption{Comparison of time and memory requirements with 3D implicit functions.}
	\begin{center}
\resizebox{\linewidth}{!}{
        \begin{tabular}{|c|c|c|c|}  
     \hline
          & OccNet~\cite{MeschederNetworks} &DISN~\cite{xu2019disn}& Ours \\   
     \hline
       \begin{tabular}{@{}c@{}}Network\\evaluations\end{tabular} & $\mathcal{O}(3.8*R^2)$ & $\mathcal{O}(R^3)$ & $\mathcal{O}(R^2)$ \\
       \hline
       Time(CPU) & 55.80s & 14.68s & \textbf{8.79}s \\
       \hline
      Space & 1175 MB & $>11$GB & \textbf{286}MB \\
     \hline
   \end{tabular}}
	\end{center}
   \label{table:time}
\end{table}

\section{Conclusion}
We propose SeqXY2SeqZ to learn the structure of 3D shapes using a discriminative neural network not only benefiting from the regularity inherent in voxel grids during both training and testing, but also avoiding cubic complexity for high memory efficiency. SeqXY2SeqZ successfully resolves the issue of dense and irregular sampling during structure learning or inference required by 3D implicit function-based methods, which leads to higher inference times compared to our approach.
This is achieved based on the encoding of voxel grids by our 1D voxel tubelization, which effectively represents a voxel grid as a mapping from discrete 2D coordinates to sequences of discrete 1D locations. This mapping further enables SeqXY2SeqZ to effectively learn the 3D structures as 2D functions. We demonstrate that SeqXY2SeqZ outperforms the state-of-the-art methods under widely used benchmarks.

{\small
\bibliographystyle{ieee_fullname}
\bibliography{paper}
}

\end{document}